\documentclass[twocolumn]{article}

\usepackage[margin=0.7in]{geometry}

\usepackage{algorithm}
\usepackage{algorithmic}
\usepackage{titling}
\usepackage{microtype}
\usepackage{graphicx}
\usepackage{subfigure}
\usepackage{booktabs} 
\usepackage[utf8]{inputenc}
\usepackage{tikz}
\usepackage{authblk}

\begin{document}

\title{\bf{Fast Task-Aware Architecture Inference}}
\author[1]{\bf{Efi Kokiopoulou}\thanks{\{efi,ahauth,sbaiz,agesmundo,bartok,jberent\}@google.com}}
\author[1]{\bf{Anja Hauth}}
\author[1]{\bf{Luciano Sbaiz}}
\author[1]{\bf{Andrea Gesmundo}}
\author[1]{\bf{Gabor Bartok}}
\author[1]{\bf{Jesse Berent}}
\affil[1]{Google AI Perception}

\renewcommand\Authands{ and }

\maketitle






\begin{abstract}
Neural architecture search has been shown to hold great promise towards the automation of deep learning. However in spite of its potential, neural architecture search remains quite costly. To this point, we propose a novel gradient-based framework for efficient architecture search by sharing information across several tasks. We start by training many model architectures on several related (training) tasks. When a new unseen task is presented, the framework performs architecture inference in order to quickly identify a good candidate architecture, \emph{before} any model is trained on the new task. At the core of our framework lies a deep value network that can predict the performance of input architectures on a task by utilizing task meta-features and the previous model training experiments performed on related tasks. We adopt a continuous parametrization of the model architecture which allows for efficient gradient-based optimization. Given a new task, an effective architecture is quickly identified by maximizing the estimated performance with respect to the model architecture parameters with simple gradient ascent. It is key to point out that our goal is to achieve reasonable performance at the lowest cost.
We provide experimental results showing the effectiveness of the framework despite its high computational efficiency.
\end{abstract}

\begin{figure}[t]
    \centering
    \subfigure[]
    {
        \includegraphics[scale=.43]{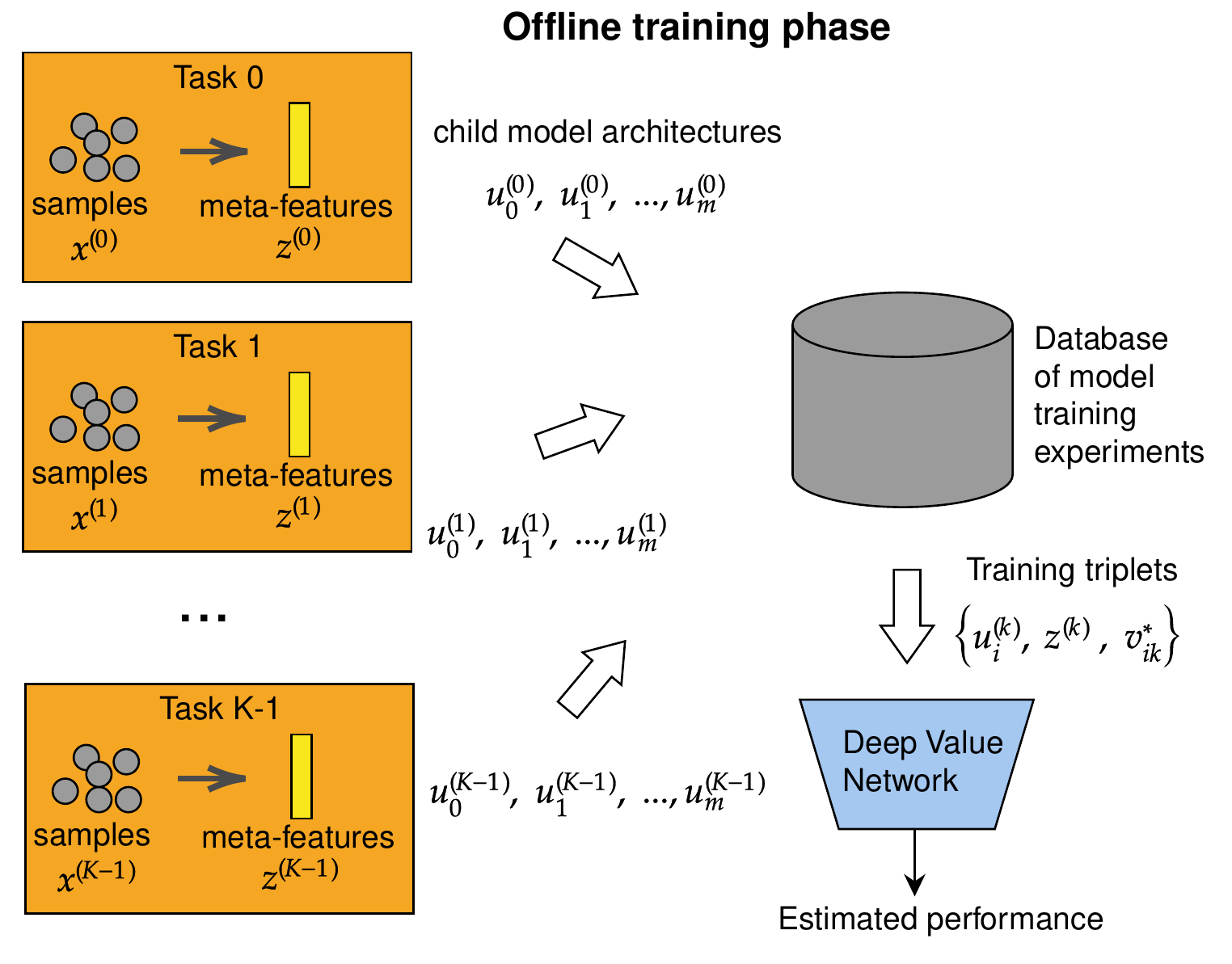}
    }
    \\
    \subfigure[]
    {
        \includegraphics[scale=.43]{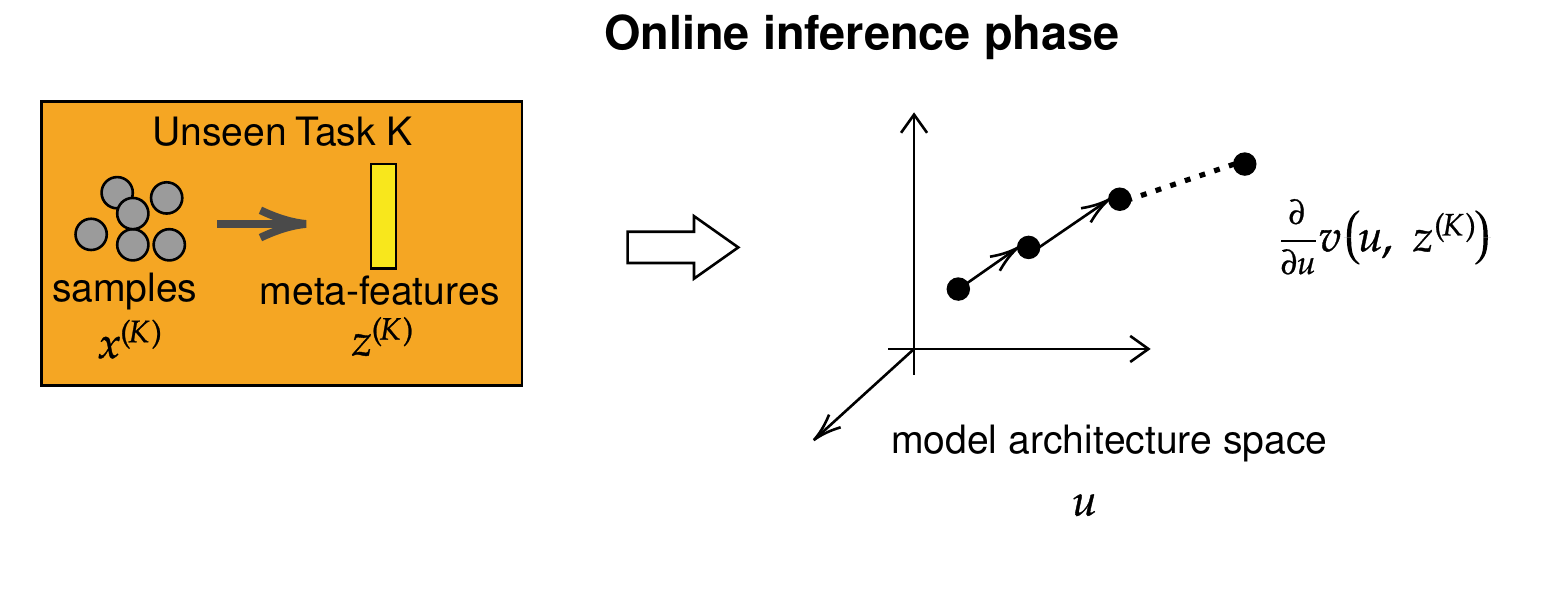}
    }
    \caption
    { The proposed framework.
        (a) Offline  phase. Several architectures are trained and their performances are stored in a database. The performances along with meta-features about the task are used to train a  value network which estimates the  performance.
        (b) Online  phase. Given a new task and its meta-features, the system
        applies gradient ascent on the output of the  value net.
    }
    \label{fig:system_overview}
\end{figure}

\section{Introduction}
Designing high performing neural networks is a time consuming task that typically requires substantial human effort. In the past few years, neural architecture search and algorithmic solutions to model building have received growing research interest as they can automate the manual process of model design. Although they offer impressive results that compete with human-designed models~\cite{ZophLe16}, neural architecture search requires large amount of computational resources for each new task. For this reason, recent methods have been proposed that focus on reducing its cost (see e.g.,~\cite{DARTS, ENAS}). This very fact becomes a major limitation in those setups that impose strict resource constraints for model design. For example, in cloud machine learning services, the client uploads a new data set and an effective model should ideally be auto-designed in minutes (or seconds). In such settings architecture search has to be very efficient, which is the main motivation for this work.

At the same time, applying independently automated model building methods to each new task requires a lot of models to be trained as well as learning how to generate high performing models from scratch. Such an approach requires a formidable amount of computational resources and is far from being scalable. On the other hand, human experts can design state-of-the-art models using prior knowledge about how existing architectures perform across different data sets. Similar to human experts, we aim to cross learn from several task data sets and leverage prior knowledge.

In this paper, we present a framework that amortizes the cost of architecture search across several tasks and remains effective thanks to the knowledge transfer between tasks. Architecture search aims at learning a mapping from a data set to a high performing architecture. We propose to formulate architecture search as a \emph{structured prediction} problem and build on top of previously proposed deep value networks~\cite{GygliNorouziAngelova17}. Given a candidate model architecture and meta-features about the task, a deep value network provides a differentiable mapping whose output estimates the performance of the input architecture on the task data set. We also adopt a continuous parametrization of the model architecture which allows for efficient gradient-based optimization of the estimated performance. Also, in contrast to previous work, e.g., \cite{FeKlEgSpBlHu15} that uses pre-computed meta-features for the task, we present a solution for learning the meta-features directly from the raw task samples as part of the deep value network weights.

The framework consists of an \emph{offline training phase} and \emph{an online inference phase} (see Fig.~\ref{fig:system_overview} for a conceptual illustration).
Assuming that we have trained several model architectures on several related (training) tasks, when a new unseen task is presented, the framework performs fast architecture optimization in order to quickly identify a good candidate architecture, \emph{before} any model training is performed. In particular, the best candidate architecture is efficiently identified by maximizing the estimated performance with respect to the model architecture parameters with simple gradient ascent.
In summary, the paper contributions are the following:
\begin{itemize}
    \item Efficient architecture search using gradient-based architecture optimization.
    \item Ability to learn the task meta-features directly from the raw task data samples.
    \item Cross learning across many tasks (by leveraging information about how various architectures perform across many tasks data sets).
\end{itemize}
We provide experimental results showing the potential of the proposed framework.
The rest of the paper is organized as follows. Section~\ref{sec:problem_formulation} formally defines the problem we are interested in. Next, in Section~\ref{sec:framework}, we introduce the proposed framework and present it in details. Section~\ref{sec:related_work} reviews related methods from the literature.
We present experimental results in Section~\ref{sec:experiments} and the conclusions and future work in Section~\ref{sec:conclusion}.

\section{Problem formulation}
\label{sec:problem_formulation}
We are interested in task-aware \emph{efficient} neural architecture search. Given a new (unseen) task data set, we would like to identify quickly an effective model architecture \emph{before} any model is trained.
We want to learn across datasets in order to amortize the cost of neural architecture search.
In particular, we want to \emph{collectively} learn from all the model training experiments and leverage this wealth of information. Instead of performing architecture search independently for each new data set, we would like to transfer the knowledge obtained from past training experiments on related tasks.
In summary, the proposed framework should have the following properties:
\begin{itemize}
\item High scalability in terms of computing resources.
\item Ability to scale and learn collectively across task data sets.
\item Ability to propose a good architecture for a new related task without training any model.
\end{itemize}

In the next section we propose a general framework that has these desired properties.

\section{Proposed framework}
\label{sec:framework}

\begin{figure}
    \centering
    \subfigure[]
    {
        \includegraphics[scale=.5]{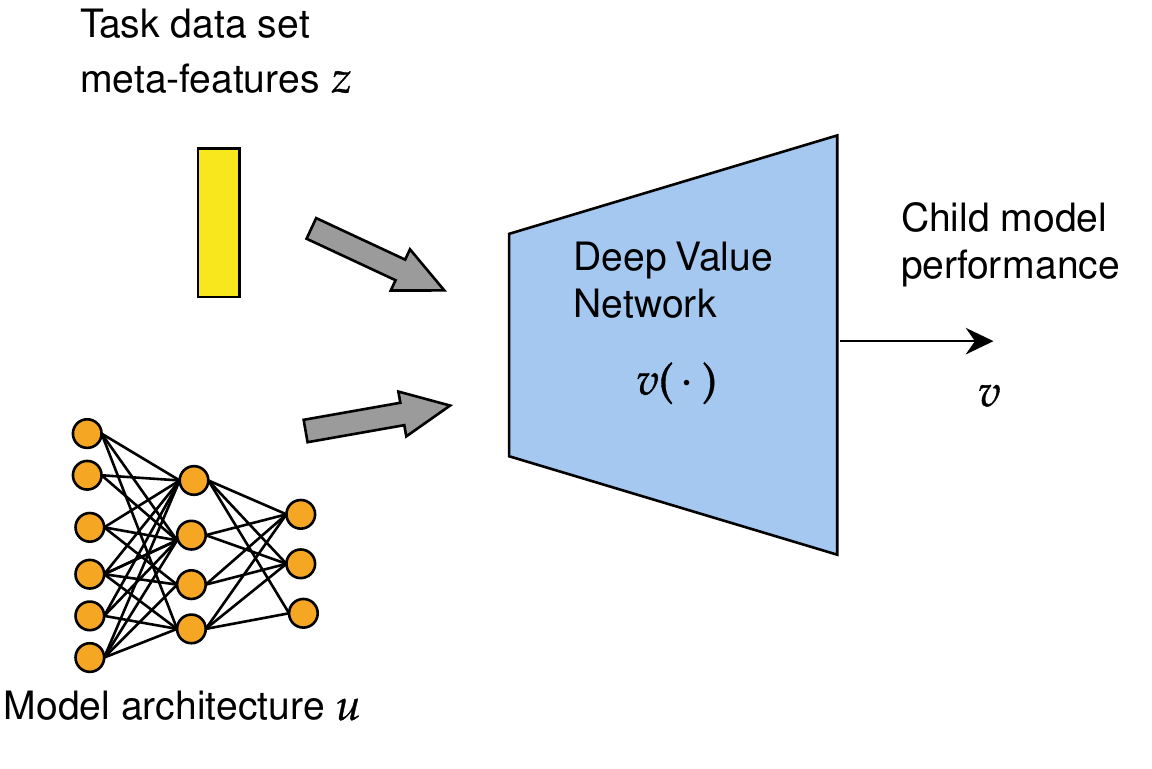}
    }
    \\
    \subfigure[]
    {
        \includegraphics[scale=.5]{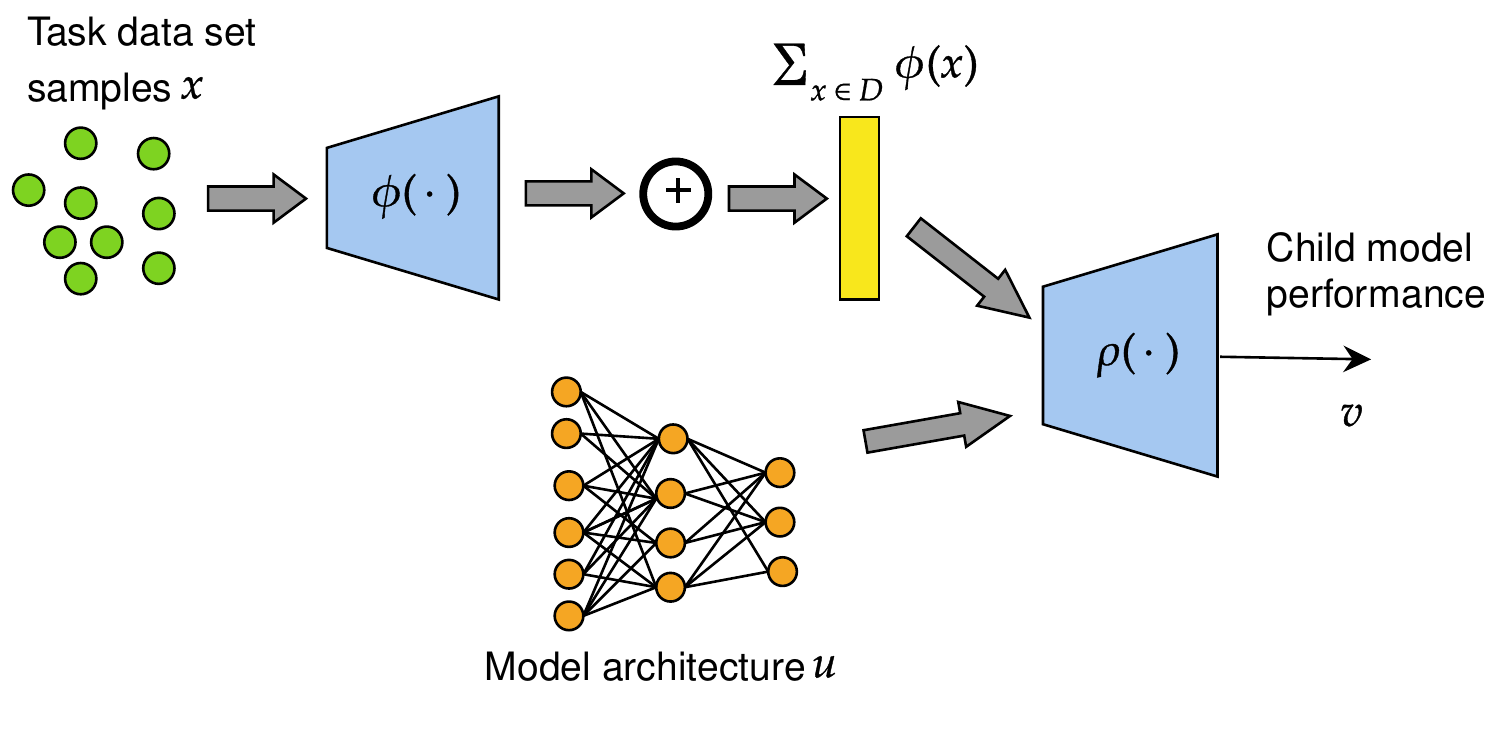}
    }
    \caption
    {Deep value network.
        (a) With pre-computed meta-features.
        (b) With learned meta-features. The value net consists of $\phi(\cdot)$ and $\rho(\cdot)$.
    }
    \label{fig:dvn}
\end{figure}

We want to automatically discover the model architecture that achieves the best quality for a given data set.
Essentially we are looking for learning a mapping from an input data set to a high performing model architecture. We propose to formalize the architecture search problem as a \emph{structured output prediction} problem. The key intuition is that learning to criticize candidate architectures is easier than learning to directly predict the optimal architecture.
In particular, the proposed framework builds on top of the \emph{Deep Value Networks} (DVNs)~\cite{GygliNorouziAngelova17} that were originally developed in the context of structured prediction applied to image segmentation. In our context, a deep value network acts as a meta-model that tunes the architecture of a child model. We consider child model families parametrized by $u$, assuming for now that $u$ is a vector of continuous variables.

A deep value network in our context takes as input: (i) descriptive meta-features $z$ derived from a certain task data set and (ii) the child model architecture parameters $u$,
and predicts how well the architecture $u$ performs on the task data set described by $z$. The performance metric $v$ can take various forms (e.g., accuracy, AUC) but the framework is agnostic to it. In this paper, we use the validation accuracy as performance metric.
The deep value network is shown conceptually in Fig. \ref{fig:dvn}. When training the value network, our hope is that it learns which type of child model architectures work well on certain types of data. This tries to mimic the human expert during manual architecture design. Human experts rely on intuition and prior knowledge when developing new candidate architectures. Here, our hope is that such an `intuition’ is encoded in the weights of the value network and that it is generally applicable and transferable across data sets. 

In the following sections, we provide more details about the proposed framework. We discuss the meta-features of a task in Section~\ref{sec:meta-features}. The framework has two phases: an \emph{off-line training} phase and an \emph{online inference} phase that are detailed below in Sections~\ref{sec:offline_phase} and~\ref{sec:online_phase} respectively. Section~\ref{sec:parametrization} discusses the child model architecture parameters $u$.

\subsection{The meta-features of a task}\label{sec:meta-features}

The meta-features $z$ of a task describe its characteristics and statistics, and they are typically derived from the task data set itself. The meta-features may include the following: total number of samples, number of classes and their distribution, label entropy, total number of features and statistics about them (min, max, median), mutual information of the features with the label and task id. The latter can be used to learn an embedding for each data set. Similar data sets should get similar embeddings (see e.g.,~\cite{WongHoulsbyLuGesmundo18}).

\paragraph{Learning the meta-features.}
On top of using pre-computed meta-features such as those listed above,
one can also learn them directly from the task data set $D$. In this case, the data set (or a large fraction of it) is given as input to the deep value network, and a task embedding is learned directly from the raw task data set samples. Note that we use both the features and the labels of the task data set samples towards learning the task embedding.
This task embedding plays the role of the meta-features and is learned jointly together with the rest of the weights of the deep value network.

The task embedding should be invariant to the order of the samples in the task data set. According to~\cite{DeepSets17}, such a function can be decomposed in the form $\rho(\sum_{x \in D} \phi(x))$ for suitable transformations $\phi$ and $\rho$. 
The latter transformations are typically implemented by a few layers (e.g., fully connected, non-linearities etc.).
The main idea is to transform each sample from the task data set using $\phi(\cdot)$ and then aggregate the transformed samples such that the task embedding becomes permutation invariant before it is fed into $\rho(\cdot)$.
This process is shown conceptually in Fig.~\ref{fig:dvn}, where the deep value network essentially consists of $\phi(\cdot)$ and $\rho(\cdot)$ that are jointly learned, i.e.,
\begin{equation}
v(u, z) := \rho\left(u, \sum_{x \in D} \phi(x)\right)
\end{equation}
We assume here that the data samples of different tasks are expressed in a common feature space that can be ingested by $\phi(\cdot)$.

\subsection{Off-line training phase}\label{sec:offline_phase}

\begin{algorithm}[tb]
   \caption{Offline training phase}
   \label{alg:offline_phase}
\begin{algorithmic}
   \STATE {\bfseries Inputs:} \\
   ~~Task datasets $D_k$, see Eq. (\ref{eq:task_data_set}) \\
   ~~DVN training set $T$, see Eq. (\ref{eq:dvn_data_set}) \\
   ~~\texttt{kInnerIters}, \texttt{kOuterIters}
   \REPEAT
   \STATE Sample randomly a task $k$.
   \FOR{$i=1$ {\bfseries to} \texttt{kOuterIters} }
   \STATE Pick a mini-batch from $T$ with samples only from task $k$.
   \FOR{$j=1$ {\bfseries to} \texttt{kInnerIters} }
   \STATE Pick a large batch from the task dataset $D_k$.
   \STATE Perform one step of Stochastic Gradient Descent step on the weights of the deep value network.
   \ENDFOR
   \ENDFOR
   \UNTIL{Convergence}
\end{algorithmic}
\end{algorithm}

Assume we have $K$ tasks with corresponding data sets denoted:
\begin{equation}\label{eq:task_data_set}
D_k = \left\{ (x_i^{(k)}, y_i^{(k)}) \right\}_{i=0}^{N_k-1}, ~ k=0,\ldots, K-1
\end{equation}
where $N_k$ is the number of data samples in the $k$-th task.
$(x_i^{(k)}, y_i^{(k)})$ is the i-the sample and its corresponding label in the $k$-th task data set.

For each task data set, we generate $m$ child model architectures, train them and collect the model performances on the validation set in a life-long database of model training experiments (see Fig. \ref{fig:system_overview}). This database is used to generate the training set for the deep value network, which consists of $M$ triplets of the form:
\begin{equation}\label{eq:dvn_data_set}
T = \left\{ (z_i, u_i, v^*_i) \right\}_{i=0}^{M-1},
\end{equation}
where the value $v^*_i$ holds the child model performance obtained when training with the model architecture $u_i$ on the task data set with meta-features $z_i$. In this paper, the model performance metric used is the validation accuracy. 
As more tasks are ingested in the database and more models get trained, the deep value network improves its predictions. In the Appendix we provide experimental results demonstrating this behaviour.

Once the child model training experiments have been collected in the database, we can start training the deep value network. Algorithm~\ref{alg:offline_phase} shows the main steps of this offline training phase.

\subsection{Online inference phase}\label{sec:online_phase}

\begin{algorithm}[tb]
   \caption{Online inference phase}
   \label{alg:online_phase}
\begin{algorithmic}
   \STATE {\bfseries Inputs:} \\
   ~~New Task dataset $D_K = \{ (x_i^{(K)}, y_i^{(K)}) \}_{i=0}^{N_K - 1}$ \\
   ~~Trained DVN model $v(u, z; w)$ \\
    ~~\texttt{kNumStartingPoints}, \texttt{kMaxIters}
   \STATE Compute the meta-features $z = \sum_{x \in D_K} \phi(x)$
   \STATE Form an empty set $S = \{ \}$ of solutions
    \FOR{$i=1$ {\bfseries to} \texttt{kNumStartingPoints} }
   \STATE Pick an initial guess $u_i^{(0)}$,  $t = 0$
   \REPEAT
   \STATE $u_i^{(t+1)}= u_i^{(t)} + \eta \frac{\partial}{\partial u} v(u_i^{(t)}, z;w)$
   \STATE $t := t + 1$
   \UNTIL{Convergence} (or $t > $ \texttt{kMaxIters})
   \STATE $ S := S ~ \cup ~ \{(\hat{v}_i, \hat{u}_i) \}$ where $\hat{u}_i$ is the found solution and $\hat{v_i}$ its corresponding value.
   \ENDFOR
   \STATE Output: $\arg \max_{(v,u) \in S} v(u)$.
\end{algorithmic}
\end{algorithm}

After training the deep value network $v(u, z; w)$, the model weights $w$ are kept fixed. At inference time, given a new task dataset, we first extract its meta-features $z$. At this point we can employ the value network in two ways. First, if we have a candidate architecture $u$ we can evaluate it by simply doing a forward pass on the deep value net and get the estimated child model performance.
Alternatively, we can compute the gradient of $v(u, z)$ with respect to $u$ and perform simple gradient-based optimization to get a good candidate architecture $\hat{u}$ that maximizes the estimated child model performance.

In practice, we noticed that the gradient ascent is sensitive to initialization. Hence, we run the process several times with different initial guesses and at the end pick the one that resulted in the maximum estimated performance.
Note also that in order to be able to perform gradient-based inference we need to relax the model architecture parameters $u$ to live in a continuous space. Section~\ref{sec:parametrization} below discusses this parametrization in details. The main steps of the online phase are shown in Algorithm~\ref{alg:online_phase}. This online process is also illustrated conceptually in Fig.~\ref{fig:system_overview}.

\subsection{Architecture parametrization}\label{sec:parametrization}
We discuss in this section the parametrization of the child model architectures. 
Previous work \cite{DARTS, ShinPackerSong2018} has shown that relaxing the parametrization from discrete to continuous space allows for efficient gradient-based optimization schemes while still providing competitive model performances. Our approach goes along the lines of this previous work. The main idea is that in order to make the architecture space continuous we move away from the categorical nature of design choices to a parametrized softmax over all possible choices.
We provide below a few examples where this is applied.

\paragraph{Continuous parametrization for one layer}
Assume that we have implemented a basis set consisting of $p$ base layers $o_i(x)$ corresponding to different sizes and different activation functions. We associate a weight $\alpha_i$ with each base layer and we define a new parametrized layer as follows
\begin{equation}\label{eq:layer_parametrization}
 o(x) = \sum_{i=1}^{p} \frac{\exp(\alpha_i)}{\sum_{j=1}^{p} \exp(\alpha_j)} o_i(x).  
\end{equation}
The values $\alpha_i$ allow the final parametrized layer $o(x)$ to `morph' from one size to another and/or from one activation function to another.
We use zero padding whenever needed to resolve the dimension mismatch among base layers of different sizes.

\paragraph{Continuous parametrization for a child network}
Leveraging on the continuous parametrization for one layer introduced above, we can  put several parametrized layers together. We attach a superscript to the layer parameters to denote the layer where they belong to i.e., $\alpha_i^{(j)}$ is the parameter that multiplies the output of the $i$-th base layer in the $j$-th parametrized layer of the final network. 
We also add the ability for each layer to be enabled or disabled independently from the other layers. For this, we add extra parameters $\beta_j$ that control the presence or absence of each layer. This is shown conceptually in Fig.~\ref{fig:parametrization} below.

\begin{figure}[h]
\centering
\includegraphics[scale=.46]{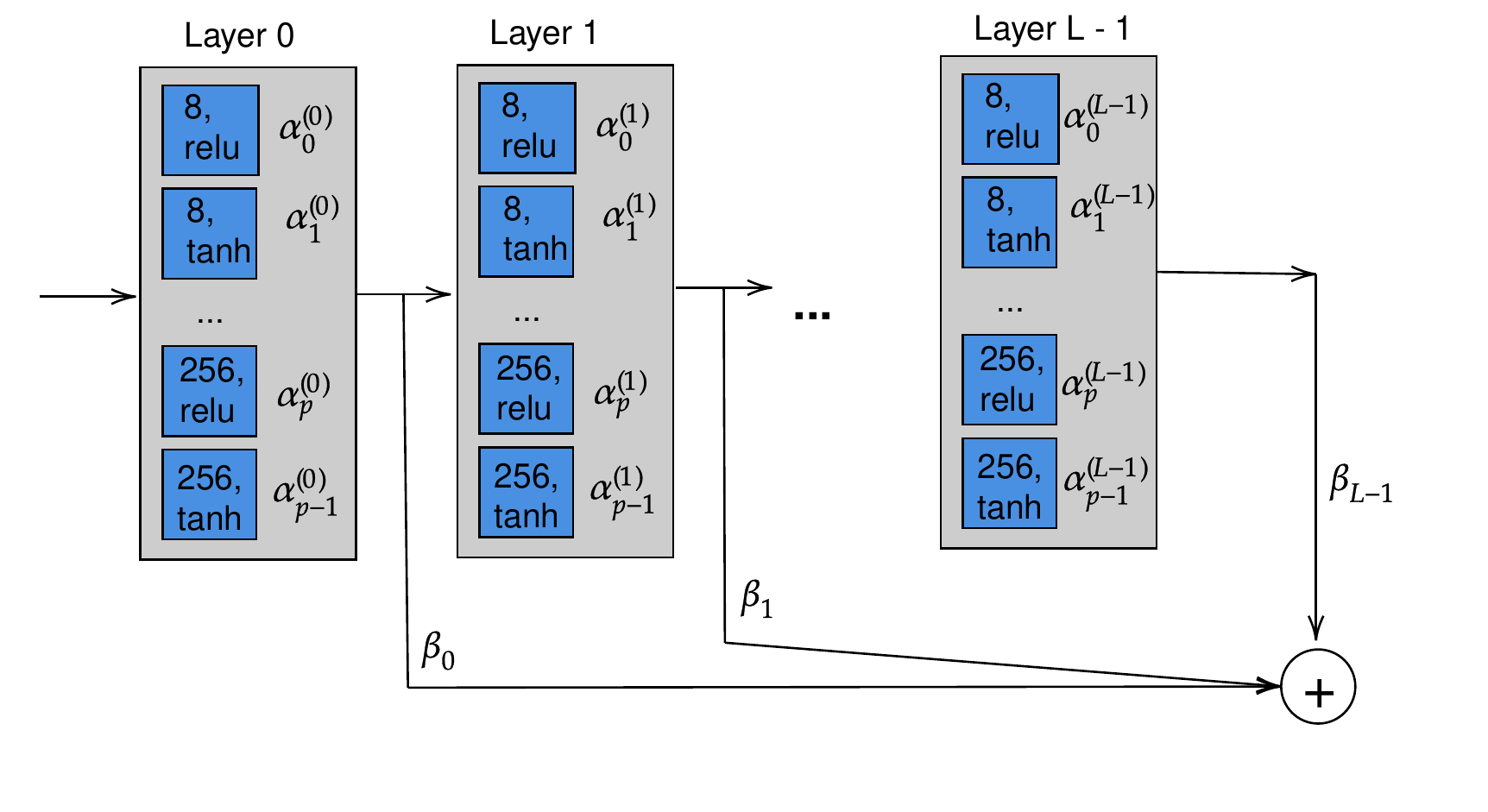}
\caption{Continuous parametrization of the child models.}
\label{fig:parametrization}
\end{figure}

Putting everything together, we consider child models that are standard Feedforward Neural Networks (FFNNs) composed of an embedding module followed by several  parametrized layers and a final softmax classification layer.  The reason for using an embedding module is that it speeds up the training time for the child models and improves their quality especially when the training set is small. The embedding module is soft-selected by an input set of pre-trained embedding modules\footnote{The pre-trained modules are available via the Tensorflow Hub service (https://www.tensorflow.org/hub). Please refer to the Appendix for more details about them.} using the same softmax trick analogous to Eq.~(\ref{eq:layer_parametrization}) where we denote by $\gamma$ the corresponding parameters of the softmax. 
After this relaxation, architecture search reduces to learning the continuous variables $u := \{ \{\alpha \}, \{\beta \}, \{\gamma \} \}$. We refer to $u$ as the encoding of the model architecture.
Finally, we would like to emphasize that this parametrization is just one example among many possible options. Any parametrization will work with the proposed framework as long as it is continuous.


\section{Related work}
\label{sec:related_work}

Automated model building is an important challenging research problem and several related methods have been proposed in the past few years.
In general, previous works can be broadly categorized into the following classes:
\begin{itemize}
\item Bayesian optimization methods~\cite{SnoekLarochelleAdams12, MeKlFeuSpHut16, SMAC2011, FuSheEli18} build a probabilistic model of the performance of the network as a function of its hyperparameters and then decide which candidate point in the search space to evaluate next.
\item Methods based on Reinforcement Learning (RL)~\cite{ZophLe16, WongHoulsbyLuGesmundo18, ENAS, PNAS} evaluate candidate child model architectures on-the-fly, by training and evaluating on a validation set. Using the validation accuracy as the reward signal, these methods use RL to optimize the child model architecture.
\item Evolutionary methods such as \cite{ReMoSeSaSuTaLeKu17} form a population of model architectures. The population is evolved over time by picking individuals and mutating them (e.g., inserting a new layer). The quality of the population improves over time as the individuals with poor performance are removed.
\item Morphing methods~\cite{MorphNet, AdaNet} start with an initial model architecture and they iteratively refine the architecture during training until a certain objective is hit (e.g., model size or flops per inference).
\item Performance prediction methods~\cite{TAPAS, Peephole, BaGuRaNa17, PNAS}. Given a candidate model architecture, these methods forecast its performance without training. In order to train the performance predictor a database of previous trainings of various model architectures is typically built.
\end{itemize}

The proposed framework also belongs to the last category of performance prediction methods. However, our deep value network is task-aware and takes-in not only the architecture but also meta-features about the task, with the extra ability of learning them directly from the raw task samples. Hence, the proposed framework in its current form is novel (to the best of our knowledge).
However, it shares connections and similarities with existing works that we outline below.
The previously proposed SMAC method \cite{SMAC2011} for general algorithm configuration also uses a history of past configuration experiments as well as descriptive features for the problem instances. However, this method uses an expensive Bayesian optimization process as opposed to the efficient gradient-based architecture search that this framework proposes (implied by the structured prediction formulation).

The TAPAS system proposed in \cite{TAPAS} also uses a history of past configuration experiments stored in a database of experiments. The paper proposes a performance predictor that takes as input the difficulty of the dataset as well as a candidate network architecture. However, they use only pre-computed meta-features and their architecture parametrization is not differentiable.

The paper in \cite{WongHoulsbyLuGesmundo18} proposes a multi-task training of RL-based architecture search methods. For each task, it learns a task embedding that captures the task similarity. The task embedding is provided as input to the controller at each time step. In contrast to our work where the task embedding is derived directly from the data samples of the task, the task embedding in \cite{WongHoulsbyLuGesmundo18} is derived from the task id. Also, this method still requires some child model trainings and evaluations on the test task as opposed to our method that requires no child model trainings.

\begin{table*}
\caption{Statistics for the NLP classification data sets. Number of examples in the training set, validation set and test set, number of classes and reference.}
\label{NLP-datasets-table}
\vskip 0.15in
\begin{center}
\begin{small}
\begin{tabular}{lccccc}
\toprule
\sc{Data set} & \sc{Train Examples} & \sc{Val. Examples} & \sc{Test Examples} & \sc{Classes} & \sc{Reference} \\
\midrule
\sc{Airline}                 & 11712 & 1464 & 1464 & 3  & crowdflower.com \\
\sc{Corporate Messaging}     & 2494  & 312 & 312   & 4  & crowdflower.com \\
\sc{Emotion}                 & 32000 & 4000 & 4000 & 13 & crowdflower.com   \\
\sc{Disasters}               & 8688  & 1086 & 1086 & 2  & crowdflower.com  \\
\sc{Global Warming}          & 3380  & 422  & 423  & 2  & crowdflower.com \\
\sc{Political Bias}          & 4000  & 500  & 500  & 2  & crowdflower.com     \\
\sc{Political Message}       & 5000  & 500  & 500  & 9  & crowdflower.com  \\
\sc{Progressive Opinion}     & 927   & 116  & 116  & 3  & crowdflower.com \\
\sc{Progressive Stance}      & 927   & 116  & 116  & 4  & crowdflower.com    \\
\sc{US Economic Performance} & 3961  & 495  & 496  & 2  & crowdflower.com \\
\bottomrule
\end{tabular}
\end{small}
\end{center}
\vskip -0.1in
\end{table*}

\section{Experiments}
\label{sec:experiments}

The framework has been implemented in TensorFlow\footnote{We plan to make the code publicly available.}~\cite{TF}.
For the experimental results we use publicly available NLP data sets whose main characteristics are shown in Table~\ref{NLP-datasets-table}.
We have performed several leave-one-out experiments, where each task in our set is considered to be a test task and  the rest of the tasks are used as the training tasks. Then for each such leave-one-out experiment, we train a DVN, we study its predictive performance and use it for fast architecture inference. More details are provided below.

\subsection{Setup}
\paragraph{Child models}

The child models have been implemented using the parametrization discussed in Section~\ref{sec:parametrization}. 
The sizes of the base layers in a single parametrized layer are $\{8, 16, 32, 64, 128, 256\}$ and each one of them is combined with two distinct activation functions (\texttt{relu} and \texttt{tanh}). Hence a single parametrized layer is composed of twelve base layers and each child model has seven such parametrized layers.
The child models have been trained using Adam optimizer~\cite{Adam} with a learning rate of $10^{-4}$ for 20 training epochs.

\paragraph{Deep value network}
The value network was trained on the child model training experiments stored in the database, which was populated with about 500 child model architectures per task (generated by random one-hot architecture encodings).
We used a simple value network consisting of two fully connected layers of size 50 each for the task meta-features tower (aka $\phi(\cdot)$ in Fig. \ref{fig:dvn}) and two fully connected layers of sizes 50 and 10 for the tower that produces the final prediction (aka $\rho(\cdot)$ in Fig.~\ref{fig:dvn}).
The value network used standard L2 loss for regression and was trained using Stochastic Gradient Descent with momentum~\cite{momentum} (using 0.5 as default parameter). The learning rate was set to $10^{-4}$.
We set \texttt{kOuterIters} to 1 and \texttt{kInnerIters} to 2 in Algorithm \ref{alg:offline_phase}.
When training the value network we normalized the child performances $v_i = (v_i - \mu_k) / \sigma_k$ using the mean $\mu_k$ and standard deviation $\sigma_k$ of the population of child performances for a certain task $k$.
Each task has its own level of difficulty and we noticed that this normalization step factors out the difficulty of the task and improves the performance of the value network.


\subsection{Predicting the model performance}

\begin{table*}
  \caption{Spearman's rank correlations between the actual performances and the predicted performances provided by the value network; breakdown by task. The higher the better. The meta-features seem to help in the majority of tasks.}
  \label{table:spearman_values}
\vskip 0.15in
\begin{center}
\begin{small}
\begin{sc}
\begin{tabular}{lcc}
\toprule
Task name & Without Meta-features & With Meta-Features  \\
\midrule
airline & 0.8003 $\pm$ 0.0180 & 0.8260 $\pm$ 0.0125 \\
emotion & 0.8269 $\pm$ 0.0138 & 0.8523 $\pm$ 0.0088 \\
global warming & 0.8072 $\pm$ 0.0116 & 0.8179 $\pm$ 0.0076 \\
corporate messaging & 0.8090 $\pm$ 0.0067 & 0.8527 $\pm$ 0.0076 \\
disasters & 0.8066 $\pm$ 0.0053 & 0.7933 $\pm$ 0.0121 \\
political message & 0.4915 $\pm$ 0.0114 & 0.5078 $\pm$ 0.0091 \\
political bias & 0.5408 $\pm$ 0.0138 & 0.5210 $\pm$ 0.0111 \\
progressive opinion & 0.8164 $\pm$ 0.0130 & 0.8338 $\pm$ 0.0063 \\
progressive stance & 0.7244 $\pm$ 0.0212 & 0.7883 $\pm$ 0.0200 \\
us economic performance & 0.3051 $\pm$ 0.0103 & 0.2851 $\pm$ 0.0114 \\
\bottomrule
\end{tabular}
\end{sc}
\end{small}
\end{center}
\vskip -0.1in
\end{table*}

\begin{table*}
  \caption{R2 values between the actual performances and the predicted performances provided by the value network; breakdown by task. The higher the better. The meta-features seem to help in the majority of tasks.}
  \label{table:R2_values}
\vskip 0.15in
\begin{center}
\begin{small}
\begin{sc}
\begin{tabular}{lcc}
\toprule
Task name & Without Meta-features & With Meta-Features  \\
\midrule
airline & 0.7709 $\pm$ 0.0143 & 0.8294 $\pm$ 0.0174 \\
emotion & 0.7570 $\pm$ 0.0164 & 0.8011 $\pm$ 0.0116 \\
global warming & 0.7002 $\pm$ 0.0102 & 0.7403 $\pm$ 0.0138 \\
corporate messaging & 0.7218 $\pm$ 0.0102 & 0.7746 $\pm$ 0.0099 \\
disasters & 0.7805 $\pm$ 0.0120 & 0.8039 $\pm$ 0.0143 \\
political message & 0.7345 $\pm$ 0.0124 & 0.7451 $\pm$ 0.0138 \\
political bias & 0.1718 $\pm$ 0.0148 & 0.1382 $\pm$ 0.0148 \\
progressive opinion & 0.4473 $\pm$ 0.0060 & 0.4720 $\pm$ 0.0092 \\
progressive stance & 0.4189 $\pm$ 0.0112 & 0.4614 $\pm$ 0.0138 \\
us economic performance & 0.6886 $\pm$ 0.0164 & 0.6970 $\pm$ 0.0146 \\
\bottomrule
\end{tabular}
\end{sc}
\end{small}
\end{center}
\vskip -0.1in
\end{table*}

We study the predictive performance of the value network in each one of the leave-one-out experiments.
In particular, given the predicted performances and their corresponding actual performances, we quantify the predictive performance in terms of the Spearman's rank correlation coefficient as well as the standard R2 metric for regression.
In order to get more accurate results we repeat this process ten times and we report the statistics of the obtained performances.
Table~\ref{table:spearman_values} shows the obtained Spearman's rank correlations for each task and Table~\ref{table:R2_values} shows the corresponding R2 metric values.
Notice that in most cases, the  Spearman's rank correlations lie around 0.8, which seems rather satisfactory for a method that does not use any child model trainings on the test task.

We have also studied experimentally the effect of the meta-features and report the predictive performances with and without meta-features. The results in both tables suggest that the meta-features are helpful, as expected, since both metrics increase. The meta-features provide task-specific information to the value network that helps towards estimating the relative performance of various architectures for the task at hand.

\subsection{Architecture search}

\begin{table*}
  \caption{Comparison with NAS in terms of test accuracy. The table shows the test accuracy achieved by the top model according to the validation accuracy that NAS found. The number of child models that NAS trained in order to achieve this test accuracy is also reported. For the sake of completeness, we report the statistics of the test accuracy obtained by the first 10 models that NAS tried. Notice that the proposed method (without any child model training on the test task) achieves test accuracy which is close to that of NAS in the majority of cases.}
  \label{table:NAS_comparison}
\vskip 0.15in
\begin{center}
\begin{small}
\begin{sc}
\begin{tabular}{lcccc}
\toprule
Task name &	First10  & NAS  & Num	Child models & Proposed 	\\
\midrule
airline & 0.7904 $\pm$ 0.0366 & 	0.83197 &	751	& 0.8222 $\pm$ 0.0129	\\
global warming & 0.7806 $\pm$ 0.0249 &	0.79196	& 1927 & 0.8017 $\pm$  0.0108	\\
disasters &	 0.8193 $\pm$ 0.0105 & 0.83425 &	1283	 &  0.8235 $\pm$ 0.0119	 \\
political bias & 0.7770 $\pm$ 0.0151 &	0.778 &	1989	&  0.7686 $\pm$ 0.0108	\\
progressive opinion &  0.6750 $\pm$ 0.0428 &	0.73276	& 1505 &	0.7052 $\pm$ 0.0381 \\	
progressive stance &  0.4181 $\pm$ 0.0645 &	0.57759	& 1635 & 0.4724 $\pm$ 0.0537	\\
us economic performance	 & 0.7494 $\pm$ 0.0112 & 0.76411	& 1966 & 0.7498 $\pm$ 0.0132 \\	
corporate messaging & 0.8006 $\pm$ 0.0492 & 0.85897 &	968	& 0.8247 $\pm$ 0.0387	\\
emotion	& 0.2998 $\pm$ 0.0278 & 0.35425	 &1779	& 0.3397 $\pm$ 0.0238	\\
political message & 0.4230 $\pm$ 0.0075 &	0.414	& 1974 &  0.4214 $\pm$ 0.0061 \\
\bottomrule
\end{tabular}
\end{sc}
\end{small}
\end{center}
\vskip -0.1in
\end{table*}

In this section we look into the performance of the child model architectures suggested by our method and we report their test accuracy. 
When we apply our algorithm we set \texttt{kNumStartingPoints} to 10 and \texttt{kMaxIters} to 1000 in Algorithm \ref{alg:online_phase}. For each task we run our method ten times in order to get more accurate statistics on the performances.

We compare against the NAS method for architecture search using Reinforcement Learning \cite{ZophLe16}.
In particular, we applied NAS on the same child models as our method.
Table~\ref{table:NAS_comparison} shows for each task the test accuracy of the child model that NAS found as having the best validation accuracy. We report also the number of trained child models that were needed for achieving this accuracy. For the sake of completeness, the table also includes the performances of the first 10 models that NAS tried.


Notice that the performance of the proposed method is not too far from that of NAS. This is very promising given that the proposed method requires no child model training in its online phase and is very efficient.

Note finally that the experiments above have been all performed in the continuous architecture space. However, we acknowledge that inference with continuous child model architectures can be expensive for some applications, since it involves computations over all possible design choices. In such cases, one may want to prune the architecture (in order to make inference faster) but still keep the same model quality. This will be the subject of a forthcoming study.



\section{Conclusions and future work}
\label{sec:conclusion}
We presented a framework for efficient architecture inference that cross learns from several tasks. This is feasible thanks to a deep value network that predicts the performance of a candidate architecture on a certain task based on learned meta-features derived from the raw data. Given a new task, the proposed method uses simple gradient ascent to infer a candidate architecture for it and experimental results confirm that the performance of the found architecture is relatively close to that of the very expensive baseline.
In our future work, we plan to explore different child model parametrizations, study the effect of pruning the architecture and apply the method to other data modalities beyond text (e.g., images).

\section*{Acknowledgements}
The authors would like to thank Thomas Deselaers for his valuable comments, fruitful discussions and support.

\bibliographystyle{plain}
\bibliography{references}

\begin{thebibliography}{10}

\bibitem{TF}
Mart\'{\i}n Abadi, Paul Barham, Jianmin Chen, Zhifeng Chen, Andy Davis, Jeffrey
  Dean, Matthieu Devin, Sanjay Ghemawat, Geoffrey Irving, Michael Isard,
  Manjunath Kudlur, Josh Levenberg, Rajat Monga, Sherry Moore, Derek~G. Murray,
  Benoit Steiner, Paul Tucker, Vijay Vasudevan, Pete Warden, Martin Wicke, Yuan
  Yu, and Xiaoqiang Zheng.
\newblock Tensorflow: A system for large-scale machine learning.
\newblock In {\em Proceedings of the 12th USENIX Conference on Operating
  Systems Design and Implementation}, OSDI'16, pages 265--283, Berkeley, CA,
  USA, 2016. USENIX Association.

\bibitem{BaGuRaNa17}
B.~Baker, O.~Gupta, R.~Raskar, and N.~Naik.
\newblock Accelerating neural architecture search using performance prediction.
\newblock {\em arXiv preprint}, November 2017.

\bibitem{AdaNet}
Corinna Cortes, Xavier Gonzalvo, Vitaly Kuznetsov, Mehryar Mohri, and Scott
  Yang.
\newblock Adanet: Adaptive structural learning of artificial neural networks.
\newblock In {\em Proceedings of the 34th International Conference on Machine
  Learning - Volume 70}, ICML'17, pages 874--883. JMLR.org, 2017.

\bibitem{Peephole}
B.~Deng, J.~Yan, and D.~Lin.
\newblock Peephole: Predicting network performance before training.
\newblock {\em arXiv preprint}, December 2017.

\bibitem{FeKlEgSpBlHu15}
M.~Feurer, A.~Klein, K.~Eggensperger, J.~Springenberg, M.~Blum, and Frank
  Hutter.
\newblock {Efficient and Robust Automated Machine Learning}.
\newblock {\em NIPS}, 2015.

\bibitem{FuSheEli18}
N.~Fusi, R.~Sheth, and H.~M. Elibol.
\newblock {Probabilistic Matrix Factorization for Automated Machine Learning}.
\newblock {\em 32nd Conference on Neural Information Processing Systems (NIPS
  2018), Montréal, Canada.}, 2018.

\bibitem{MorphNet}
A.~Gordon, E.~Eban, O.~Nachum, B.~Chen, H.~Wu, T.-J. Yang, and E.~Choi.
\newblock {MorphNet: Fast \& Simple Resource-Constrained Structure Learning of
  Deep Networks}.
\newblock {\em CVPR}, 2018.

\bibitem{GygliNorouziAngelova17}
M.~Gygli, M~Norouzi, and A.~Angelova.
\newblock Deep value networks learn to evaluate and iteratively refine
  structured outputs.
\newblock {\em ICML}, 2017.

\bibitem{SMAC2011}
F.~Hutter, H.~H. Hoos, and K.~Leyton-Brown.
\newblock {Sequential Model-Based Optimization for General Algorithm
  Configuration}.
\newblock {\em 5th International Conference on Learning and Intelligent
  Optimization}, pages 507--523, 2011.

\bibitem{TAPAS}
R.~Istrate, F.~Scheidegger, G.~Mariani, D.~Nikolopoulos, C.~Bekas, and A.~C.~I.
  Malossi.
\newblock {TAPAS: Train-less Accuracy Predictor for Architecture Search}.
\newblock {\em arXiv preprint}, 2018.

\bibitem{Adam}
Diederik~P. Kingma and Jimmy Ba.
\newblock Adam: {A} method for stochastic optimization.
\newblock {\em CoRR}, abs/1412.6980, 2014.

\bibitem{PNAS}
Chenxi Liu, Barret Zoph, Maxim Neumann, Jonathon Shlens, Wei Hua, Li-Jia Li,
  Li~Fei-Fei, Alan Yuille, Jonathan Huang, and Kevin Murphy.
\newblock Progressive neural architecture search.
\newblock In {\em The European Conference on Computer Vision (ECCV)}, September
  2018.

\bibitem{DARTS}
Hanxiao Liu, Karen Simonyan, and Yiming Yang.
\newblock {DARTS}: Differentiable architecture search.
\newblock In {\em International Conference on Learning Representations}, 2019.

\bibitem{MeKlFeuSpHut16}
H.~Mendoza, A.~Klein, M.~Feurer, J.~T. Springenberg, and F.~Hutter.
\newblock {Towards Automatically-Tuned Neural Networks}.
\newblock {\em JMLR: Workshop and Conference Proceedings}, 1:1--8, 2016.

\bibitem{ENAS}
Hieu Pham, Melody Guan, Barret Zoph, Quoc Le, and Jeff Dean.
\newblock Efficient neural architecture search via parameters sharing.
\newblock In Jennifer Dy and Andreas Krause, editors, {\em Proceedings of the
  35th International Conference on Machine Learning}, volume~80 of {\em
  Proceedings of Machine Learning Research}, pages 4095--4104. PMLR,
  Stockholmsmässan, Stockholm Sweden, 10--15 Jul 2018.

\bibitem{momentum}
Ning Qian.
\newblock On the momentum term in gradient descent learning algorithms.
\newblock {\em Neural Netw.}, 12(1):145--151, January 1999.

\bibitem{ReMoSeSaSuTaLeKu17}
E.~Real, S.~Moore, A.~Selle, S.~Saxena, Y.~L. Suematsu, J.~Tan, Q.~V. Le, and
  A.~Kurakin.
\newblock {Large-Scale Evolution of Image Classifiers}.
\newblock {\em Proceedings of the 34 th International Conference on Machine
  Learning, Sydney, Australia, PMLR}, 2017.

\bibitem{ShinPackerSong2018}
R.~Shin, C.~Packer, and D.~Song.
\newblock {Differentiable Neural Network Architecture Search}.
\newblock {\em Workshop track - ICLR}, 2018.

\bibitem{SnoekLarochelleAdams12}
Jasper Snoek, Hugo Larochelle, and Ryan~P Adams.
\newblock {Practical Bayesian Optimization of Machine Learning Algorithms}.
\newblock In F.~Pereira, C.~J.~C. Burges, L.~Bottou, and K.~Q. Weinberger,
  editors, {\em Advances in Neural Information Processing Systems 25}, pages
  2951--2959. Curran Associates, Inc., 2012.

\bibitem{WongHoulsbyLuGesmundo18}
Catherine Wong, Neil Houlsby, Yifeng Lu, and Andrea Gesmundo.
\newblock Transfer learning with neural automl.
\newblock In S.~Bengio, H.~Wallach, H.~Larochelle, K.~Grauman, N.~Cesa-Bianchi,
  and R.~Garnett, editors, {\em Advances in Neural Information Processing
  Systems 31}, pages 8366--8375. Curran Associates, Inc., 2018.

\bibitem{DeepSets17}
M.~Zaheer, S.~Kottur, S.~Ravanbakhsh, B.~Poczos, R.~Salakhutdinov, and
  A.~Smola.
\newblock {Deep Sets}.
\newblock {\em NIPS}, 2017.

\bibitem{ZophLe16}
B.~Zoph and Q.~V. Le.
\newblock {Neural Architecture Search with Reinforcement Learning}.
\newblock {\em ICLR}, 2017.

\end{thebibliography}
\end{document}


\title{\centering ``Fast Task-Aware Architecture Inference": \newline  Supplementary Material}
\maketitle

\section{Adding more training tasks improves performance}

\begin{figure}[p]
    \centering
    \subfigure[]
    {
        \includegraphics[scale=.7]{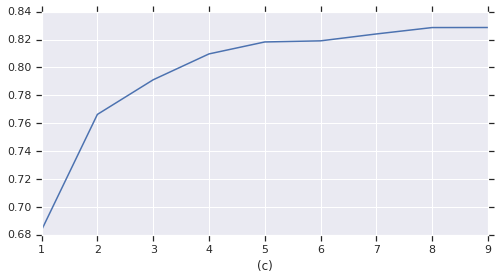}
    }
    \\
    \subfigure[]
    {
        \includegraphics[scale=.7]{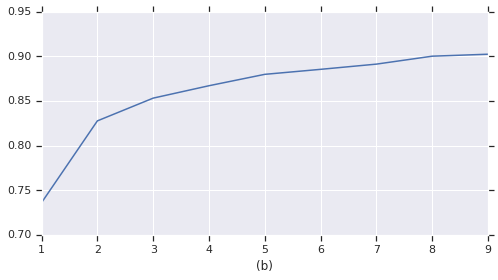}
    }
    \caption
    {The effect of adding more training tasks to the deep value net. The Spearman's rank correlation (a) and the R2 values (b) as a function of the number of training tasks. The test task here is the `airline.' Similar behaviour holds for the other tasks (omitted here).}
    \label{fig:more_tasks_in_dvn}
\end{figure}

We study the effect of gradually adding more training tasks when training the deep value network.
In order to get accurate results, we perform the experiment 100 times with different random orderings of the added tasks and for each such random ordering we train the value network 10 times (all with identical setup). In Fig. \ref{fig:more_tasks_in_dvn}, we report the average values of the Spearman's rank correlation and the R2 coefficient as a function of the number of training tasks. The experiments suggest that adding more tasks improves on average the performance of the deep value network. We noticed similar behaviour for the other tasks as well.

\section{Meta-features visualizations}

\begin{figure}[p]
\centering
\includegraphics[scale=.45]{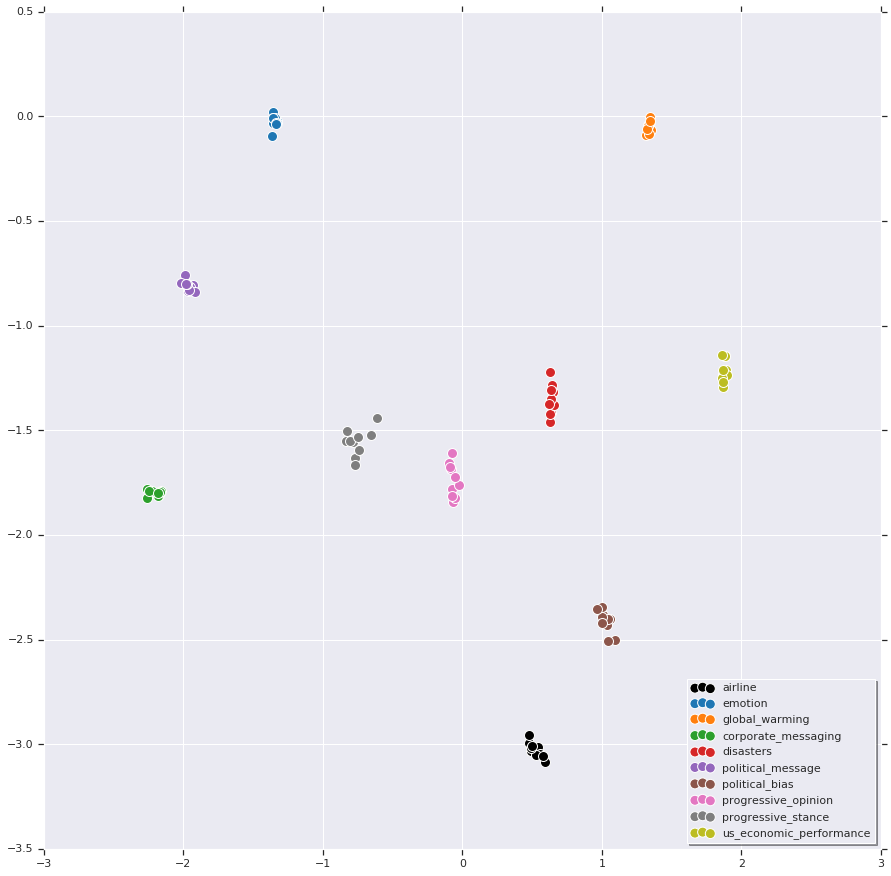}
\caption{Visualization of the learned meta-features using tSNE for the test task `airline' (black color) and the training tasks (other colors). For each task we show the meta-features computed from 10 random batches of the task samples.}
\label{fig:tSNE}
\end{figure}

\begin{figure}[p]
\centering
\includegraphics[scale=.45]{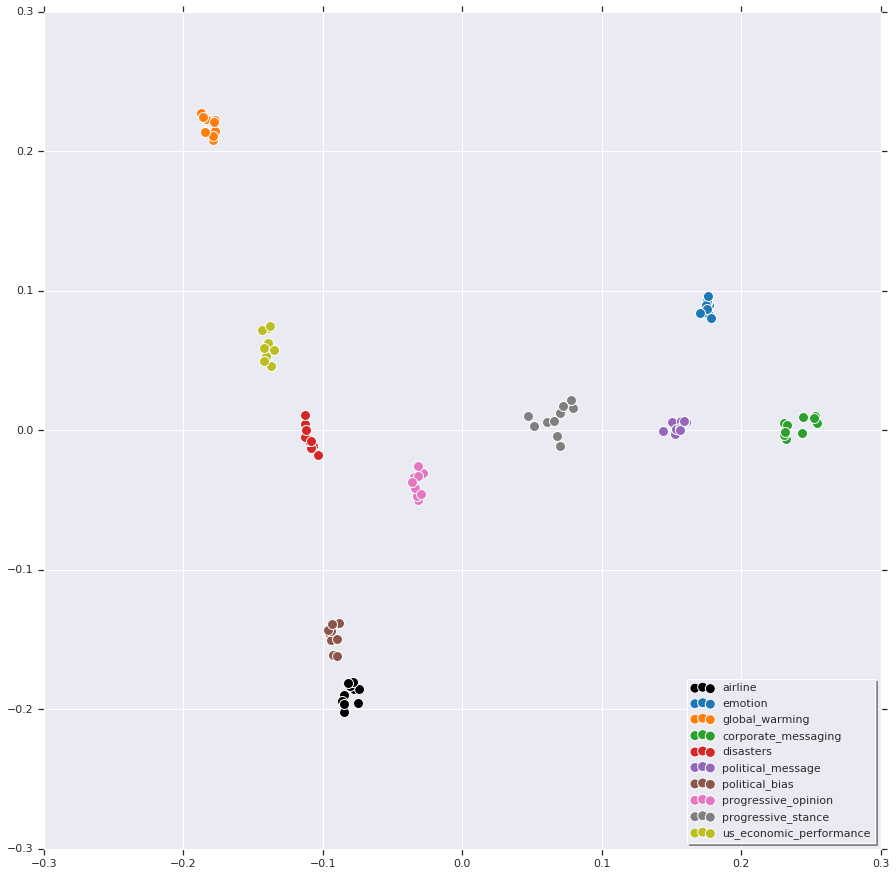}
\caption{Visualization of the learned meta-features using PCA for the same setup as in Fig.~\ref{fig:tSNE}.}
\label{fig:PCA}
\end{figure}

We looked into the learned task representations in the meta-feature space. In particular, for each task we compute
the task embedding for different batch realizations (of the task samples) and we visualized them both with tSNE as well as with Principal Component Analysis (PCA). Figures \ref{fig:tSNE} and \ref{fig:PCA} show the 2D visualizations for tSNE and PCA respectively for 10 random batches of the training and the test tasks. For tSNE we set the perplexity to 70.
One interesting observation is that different batch realizations from the same task result in close-by task embeddings in the meta-feature space, which confirms the stability of the method in this respect.

\section{Text input embedding modules}

\begin{table*}[h]
\caption{TensorFlow Hub embedding modules for text input.}
\label{table:hub_modules}
\vskip 0.15in
\begin{center}
\begin{small}
\begin{tabular}{lccl}
\toprule
Language & Dataset size & Embed dim.  & TensorFlow Hub Handles  \\
         &              &             & Prefix: \texttt{https://tfhub.dev/google/...}  \\
\midrule
English & 4B   & 250 & \href{https://tfhub.dev/google/Wiki-words-250/1}{\texttt{Wiki-words-250/1}}  \\
English & 200B & 128 & \href{https://tfhub.dev/google/nnlm-en-dim128/1}{\texttt{nnlm-en-dim128/1}}  \\
English & 7B   & 50  & \href{https://tfhub.dev/google/nnlm-en-dim50/1}{\texttt{nnlm-en-dim50/1}} \\
English & -    & 512 & \href{https://tfhub.dev/google/universal-sentence-encoder/1}{\texttt{universal-sentence-encoder/1}} \\
English & -    & 512 & \href{https://tfhub.dev/google/universal-sentence-encoder/2}{\texttt{universal-sentence-encoder/2}} \\
English & 32B  & 200 &  \\
English & 32B  & 200 &  \\
\bottomrule
\end{tabular}
\end{small}
\end{center}
\vskip -0.1in
\end{table*}

